\documentclass[10pt,twocolumn,letterpaper]{article}

\usepackage{iccv}              

\definecolor{iccvblue}{rgb}{0.21,0.49,0.74}
\usepackage[pagebackref,breaklinks,colorlinks,allcolors=iccvblue]{hyperref}
\usepackage{svg}
\usepackage{multirow}
\usepackage{graphicx}
\usepackage{placeins}
\usepackage[accsupp]{axessibility}

\def\paperID{4740} 
\def\confName{ICCV}
\def\confYear{2025}

\title{Image as an IMU: Estimating Camera Motion from a Single Motion-Blurred Image}

\author{Jerred Chen\\
University of Oxford\\
Department of Computer Science\\
{\tt\small jerred.chen@cs.ox.ac.uk}
\and
Ronald Clark\\
University of Oxford\\
Department of Computer Science\\
{\tt\small ronald.clark@cs.ox.ac.uk}
}

\begin{document}
\maketitle
\begin{abstract}
    In many robotics and VR/AR applications, fast camera motions lead to a high level of motion blur, causing existing camera pose estimation methods to fail. In this work, we propose a novel framework that leverages motion blur as a rich cue for motion estimation rather than treating it as an unwanted artifact. Our approach works by predicting a dense motion flow field and a monocular depth map directly from a single motion-blurred image. We then recover the instantaneous camera velocity by solving a linear least squares problem under the small motion assumption. In essence, our method produces an IMU-like measurement that robustly captures fast and aggressive camera movements. To train our model, we construct a large-scale dataset with realistic synthetic motion blur derived from ScanNet++v2 and further refine our model by training end-to-end on real data using our fully differentiable pipeline. Extensive evaluations on real-world benchmarks demonstrate that our method achieves state-of-the-art angular and translational velocity estimates, outperforming current methods like MASt3R and COLMAP.
\end{abstract}    

\section{Introduction}
\label{sec:intro}

\begin{figure}[t!]
  \centering
  \includegraphics[width=0.99\columnwidth]{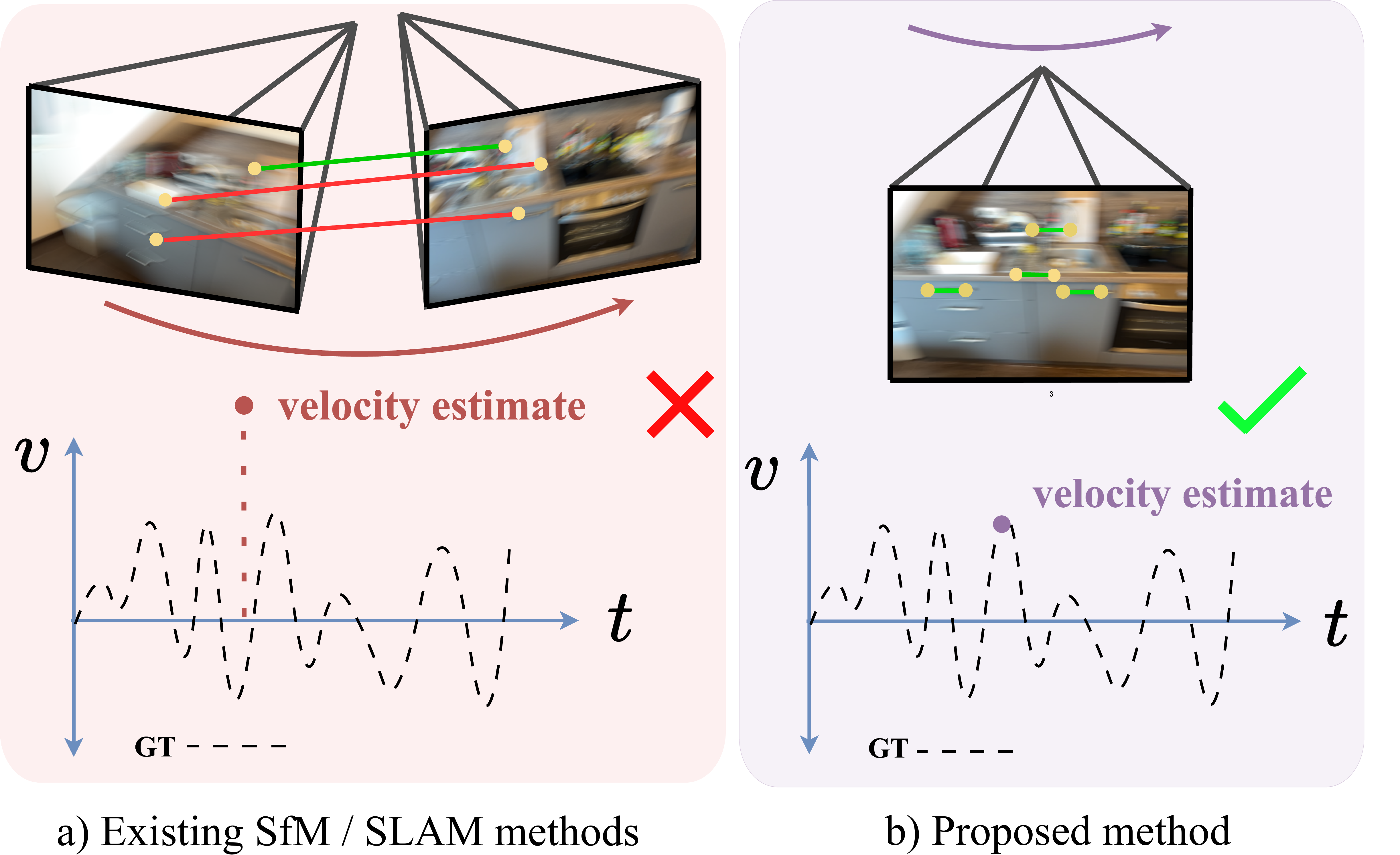}
  \caption{Existing methods rely on establishing correspondences between multiple frames to estimate inter-frame camera motion (a). This leads to failures during fast motion with motion blur. We propose a method that can estimate intra-frame motion from a single image (b), making our method robust to aggressive motions.}
  \label{fig:abstract_figure}
\end{figure}

Given a sequence of images, visual odometry (VO) and Structure from Motion (SfM) methods can accurately estimate camera poses even in moderately challenging conditions. However, these methods assume that the camera remains largely stationary during exposure, treating each image as a snapshot of the scene that can be matched against subsequent images to compute relative camera poses. This assumption, however, ignores the continuous nature of camera motion and becomes especially problematic during fast movements where motion blur can severely degrade accuracy. Conventional approaches either discard motion-blurred frames or use inertial measurement units (IMUs) to improve robustness against blur. However, using IMUs introduces additional challenges such as sensor synchronization and drift. In this work, we therefore take a fundamentally different approach: rather than treating motion blur as an unwanted artifact, we exploit it as a rich source of information about the camera's  motion (see Figure \ref{fig:abstract_figure}).

Our key insight is that the extent and direction of motion blur provide direct cues about the camera's motion. Building on this, we propose a model that, using just the motion-blur in an image, can estimate the relative motion that the camera undergoes during the exposure period. Our model works in two stages: first we predict the motion flow field and a monocular depth map, then we recover the relative pose of the camera by solving a linear least squares system. Notably, our method does not require any explicit deblurring process, and with a known exposure time, it yields an instantaneous rotational and translational velocity estimate that can be interpreted as an IMU-like measurement.

Since no existing dataset includes all the necessary data to train such a model, we construct our own dataset using a subset of ScanNet++v2 \cite{yeshwanthliu2023scannetpp} and show how we obtain realistic motion blur that generalizes to in-the-wild images. Furthermore, the fully differentiable nature of our method enables us to refine our predictions using real motion-blurred images, even when flows and depths are unavailable for training. We evaluate our model's angular and translational velocity estimates on real-world motion blurred images to test the generalization of our model. Our method runs in real-time and obtains more accurate velocity estimates compared to the existing state-of-the-art (SOTA) methods.

In summary, our contributions are as follows:
\begin{enumerate}
\item A new approach for estimating camera motion that uses the blur present in a frame to provide robust and accurate velocity estimates 
\item A pipeline for synthesizing training data (blurry images with ground-truth motion) from standard vision datasets 
\item An evaluation of our method on real-world sequences, showing our method's SOTA results and robustness to aggressive camera motion
\end{enumerate}

The rest of the paper is as follows. We first discuss relevant works to our method in Section \ref{sec:related}. We then introduce preliminaries and give high-level intuition of our method in Section \ref{sec:preliminaries}. In Section \ref{sec:method}, we describe our method in more detail. Section \ref{sec:dataset} shows the dataset construction process. We evaluate our results against baseline methods in Section \ref{sec:experiments}.
\section{Related Works}
\label{sec:related}

\textbf{Motion from Blur.}
This work is related to \textit{motion from blur} methods that attempt to extract pixel-wise flow from a single blurred image. Early approaches assumed spatially invariant blur, with \cite{chen_motionsmear_1996} using a linear motion model and \cite{rekleitis_motion_1996} using a frequency domain approach. Later methods extended these where \cite{dai_motion_2008} extracts an alpha channel matte and uses the edge gradients to estimate object motion, \cite{schoueri_optical_2009} builds on \cite{rekleitis_motion_1996} to use the color channels directly while \cite{gast_parametric_2016} proposed a variational technique and \cite{argaw_optical_2021} a learned approach. Our focus, however, is not on estimating the flow field but rather the actual camera motion.

Several works aim to recover either rotational or translational camera motion directly from a blurred image. A seminal work in this direction is \cite{Klein2005ASV} which proposes estimating the camera rotation directly from a single motion-blurred image by searching for edgels to estimate the rotation axis, which is used to compute the blur length and subsequently the rotation. \cite{li2006rot-estimate} \cite{Boracchi2008EstimatingCR} also estimate the camera rotation with assuming negligible translations, while \cite{Boracchi2009dir3d} \cite{cortes-osorio_velocity_2019} estimate exclusively linear motion. In contrast, we estimate the full 6DoF velocity. Closer to our method, \cite{zhao2023_motion-from-blur} proposes a neural network to estimate the full relative pose from a single motion-blurred image but directly regresses the pose parameters and requires a clean depth map as input.

Our method estimates depths to retrieve the metric-scaled motion. Similarly, \cite{lee_dense_2013}, \cite{liu_mba-vo_2021}, and \cite{qiu2019cvpr_worldfromblur} recover 3D geometry in motion-blurred scenes. \cite{liu_mba-vo_2021} is the most relevant, being a visual odometry method which computes the homography during exposure time between a sharp and potentially blurry frame. However, this depends on two images and an off-the-shelf deblurring method, whereas we do not require any deblurring for a single input image.

\textbf{Deblurring.}
An adjacent line of work is deblurring, which recovers a sharp image given a blurred one. Similar to our method, \cite{hu2014_jointdepth} \cite{Park2017JointEO} \cite{pan2019_single} also estimate depths and camera poses but use classical energy minimization frameworks to compute them along with the deblurred image. Compared to deep-learned deblurring methods \cite{Zhang2022DeepID}, classical methods are notably slower during test time and underperform in more general conditions. Thus, we do not benchmark against such methods given their restrictive assumptions and large runtime slowdown. Several learned deblurring methods \cite{Sun2015LearningAC} \cite{gong2017blur2mf} \cite{zhang_exposure_2022} share parallels to our work in estimating a motion flow field. Related, \cite{Wang_2022_CVPR} reformulates removing rolling-shutter effects as a deblurring problem with rolling shutter global reset. While our motion field estimates could in theory compute a deblur kernel, we aim to exploit motion cues from blur rather than remove it.

\textbf{Camera Pose Estimation.}
Camera pose estimation is a fundamental task for 3D reconstruction which can be categorized into SfM and SLAM/VO methods. SfM methods operate on unordered images and estimate the camera intrinsics and extrinsics. SLAM/VO methods can be viewed as a subset of SfM with enforced real-time constraints.

COLMAP \cite{schoenberger2016mvs} \cite{schoenberger2016sfm} is commonly the default SfM choice for estimating camera parameters and scene geometry, while ORB-SLAM \cite{orbslam3} and DSO \cite{Engel-et-al-pami2018} are classic real-time alternatives. While highly accurate in sequences with large overlap, these systems succumb to failures under conditions including sparse views and motion blur. These failures motivate learning based approaches which are more robust in these scenarios. Several fully-differentiable SfM pipelines have been proposed for greatly improving 3D reconstruction and pose estimation accuracy and robustness \cite{wang2023vggsfm} \cite{smith24flowmap} \cite{brachmann2024acezero}. In particular, DUSt3R \cite{dust3r_cvpr24} and MASt3R \cite{mast3r_arxiv24} shown how scaling pointmap regression yields SOTA performance in challenging sparse-overlap settings, inspiring a line of ``-3R''-related works \cite{zhang2024monst3r} \cite{wang2024spann3r} \cite{smart2024splatt3r} \cite{Yang_2025_Fast3R} \cite{cut3r}. SLAM and VO methods have also underwent its own deep learning revolution \cite{Czarnowski2020deepfactors} \cite{tartanvo2020corl} \cite{teed2021droid} \cite{Sucar_2021_imap} \cite{Rosinol20icra-Kimera} \cite{murai2024mast3rslamrealtimedenseslam}, with DROID-SLAM \cite{teed2021droid} obtaining SOTA pose estimates by using RAFT \cite{teed_raft_2020} for estimating optical flow. Since no prior methods estimate camera motion from a single image, we decide to benchmark against COLMAP, MASt3R, and DROID-SLAM, approximating velocities via finite differences.
\begin{figure*}[htbp!]
  \centering
    \includegraphics[width=1.0\linewidth]{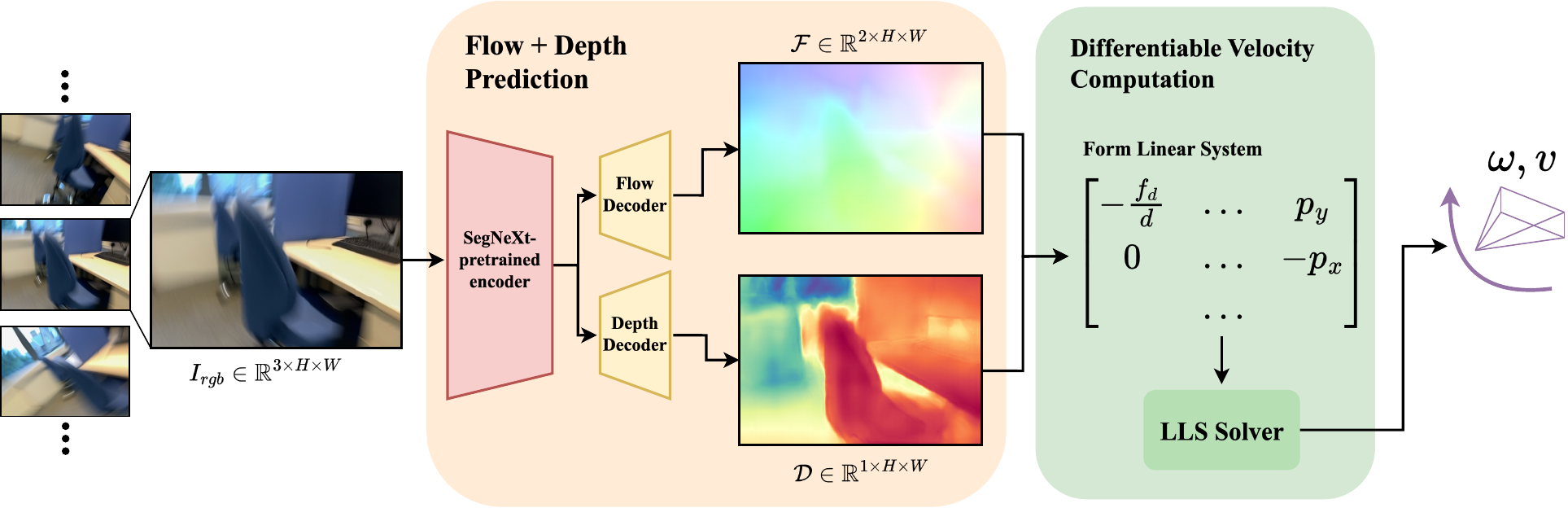}
  \caption{Method overview. Given a single motion blurred image, we pass it through the network to predict the flow field and metric depth (Section \ref{sec:flowdepth}). These are then formulated in a linear system, where the optimal velocity parameters are solved for using linear least squares (Section \ref{sec:velocity}). Because the linear solver is fully differentiable, we can train the entire network end-to-end, supervised on the camera motion.}
  \label{fig:blur_method}
\end{figure*}

\section{Preliminaries}
\label{sec:preliminaries}

Before introducing our method, we first revisit the image formation process. During the opening of the shutter, photons come into contact with the camera sensor and are collected to capture each pixel's intensity. If the camera has moved during the exposure time, each sensor pixel receives photons from different parts of the scene and motion-blur is created. The length of the motion blur traces depends on the speed of the camera as well as the exposure time; that is, when the exposure time is lengthened (e.g. in low-light settings) or when the camera moves faster relative to the scene, the blur traces become longer. More formally, let $I^B \in \mathbb{R}^{C \times H \times W}$ represent a blurry image, such that:

\begin{equation}
    I^B = g \left( \frac{1}{\tau} \int_{0}^{\tau} I^\nu(t) dt \right)
\end{equation}
\\
where $I^\nu(t) \in \mathbb{R}^{C \times H \times W}$ represents the response of the sensor to incoming photons at timestep $t$, the exposure time length denoted by $\tau$, and $g(\cdot)$ representing the linear space to sRGB conversion \cite{rim_2022_ECCV}. In practice, we approximate the blurring process in a discretized fashion:

\begin{equation}
    \label{eq:approx_blur}
    I^B \approx g \left( \frac{1}{N} \sum_{i=1}^{N} I^\nu_i \right)
\end{equation}
\\
such that $\{I^\nu_1, ..., I^\nu_N\}$ are the $N$ ``virtual images" during the exposure time.
In other words, the \textit{true} image is the average of all linear space virtual images within exposure time.

The intuition for our method is that each motion-blurred image can be seen as a composite of several virtual images, each capturing the camera at a different pose during the exposure. Each individual virtual image contributes a part of the overall blur, and the blur traces therefore act as cues that represent the motion the camera has undergone during the exposure. In essence, these blur traces can be seen as providing ``virtual correspondences" between the first and last virtual image. We hypothesize that we can extract these virtual correspondences, revealing how each point in the scene moved across the image plane and capturing information similar to an optical flow field. By assuming a rigid scene, each virtual correspondence can be interpreted as a match between two virtual images representing the camera at the beginning and end of the exposure period (see Figure \ref{fig:abstract_figure}).

Furthermore, by estimating a depth map from the blurred image, we can combine the pixel motion extracted from the motion blur traces with the scene geometry to solve for the relative pose from the start to end of the exposure. 
\section{Method}
\label{sec:method}

An overview of our method can be seen in Figure \ref{fig:blur_method}. Our approach consists of two stages. The first stage takes as input a single frame and predicts a flow field representing the motion the camera underwent during the exposure and a monocular depth map. The second stage uses the depth and flow to solve for the instantaneous velocity of the camera using a differentiable least squares solver.

\subsection{Flow and Depth Prediction}
\label{sec:flowdepth}

We follow GeoCalib \cite{veicht2024geocalib} and use SegNeXt \cite{guo2022segnext} as the backbone for our network. The blurry image $I^B$ is passed through a shared SegNeXt encoder that maps to separate decoders, outputting the pixel-wise flow field $\mathcal{F} \in \mathbb{R}^{2 \times H \times W}$ and depth map $\mathcal{D} \in \mathbb{R}^{1 \times H \times W}$. Each flow vector $\mathbf{F} = [F_x, F_y]^\top \in \mathcal{F}$ is defined to be the pixel displacement from a given pixel location in the original virtual frame $I^\nu_1$:

\begin{equation}
    \label{eq:flow_def}
    \mathbf{F} = \mathbf{p}'_2 - \mathbf{p}_1
\end{equation}
\\
such that $\mathbf{p} = [p_x, p_y]^\top$, the normalized pixel coordinates with respect to the center of the image, and $\mathbf{p}_1, \mathbf{p}'_2$ are the corresponding pixel coordinates from the first and last virtual frames $I^\nu_1$ and $I^\nu_N$ respectively.

We impose an L1 loss for the flow and depths:

\begin{equation}
    \mathcal{L}_1 = \lambda_F||\mathcal{F} - h_f(\mathcal{\hat{F}}_{fw}, \mathcal{\hat{F}}_{bw}) || + \lambda_D || \mathcal{D} - \mathcal{\hat{D}}||
\end{equation}
\\
where $\hat{\cdot}$ represents the ground truth labels, $\lambda_F, \lambda_D$ are weights to balance the losses, and $h_f$ is the reorientation function defined as:

\begin{equation}
\label{eq:reorient_flow}
h_f(\mathcal{\hat{F}}_{fw}, \mathcal{\hat{F}}_{bw} ; \mathcal{F}) = 
\begin{cases}
      \mathcal{\hat{F}}_{fw} & \text{if } \langle \mathcal{\hat{F}}_{fw}, \mathcal{F} \rangle > \langle \mathcal{\hat{F}}_{bw}, \mathcal{F} \rangle \\
      \mathcal{\hat{F}}_{bw} & \text{otherwise}
\end{cases}
\end{equation}
\\
such that $\langle \cdot, \cdot \rangle$ is the Frobenius inner product, and $\mathcal{\hat{F}}_{fw}, \mathcal{\hat{F}}_{bw}$ are the corresponding flow fields in the $I^\nu_1$ to $I^\nu_N$ direction and the $I^\nu_N$ to $I^\nu_1$ direction, respectively. Because determining the flow and depth from the true start virtual frame is ill-posed, we incorporate our reorientation function $h_f$ so that the label $\mathcal{\hat{F}}$ is closest to the predicted direction of $\mathcal{F}$. This stabilizes the training and enables the model to produce globally consistent outputs.

\subsection{Differentiable Velocity Computation}
\label{sec:velocity}

We can now use $\mathcal{F}$ and $\mathcal{D}$ to estimate the relative translation and rotation parameters $\mathbf{t} = [t_x, t_y, t_z]^\top \in \mathbb{R}^3$ and $\boldsymbol{\theta} = [\theta_x, \theta_y, \theta_z]^\top \in \mathbb{R}^3$ across exposure. To do so, we adapt the motion field equations described in Trucco and Verri \cite{trucco1998_3dcv} (see supplementary for derivation), which we express as:

\begin{equation}
\label{eq:flow_vel_x}
\begin{split}
    F_x &= \frac{t_z p_x - t_x f}{d} - \theta_y f + \theta_z p_y \\
    &+ \frac{\theta_x p_x p_y}{f} - \frac{\theta_y (p_x)^2}{f}
\end{split}
\end{equation}

\begin{equation}
\label{eq:flow_vel_y}
\begin{split}
    F_y &= \frac{t_z p_y - t_y f}{d} + \theta_x f - \theta_z p_x \\
    &- \frac{\theta_y p_x p_y}{f} + \frac{\theta_x (p_y)^2}{f}.
\end{split}
\end{equation}
\\
such that $d \in \mathcal{D}$ and $f$ is the known focal length.

These equations can be expressed in matrix form as $\mathbf{A} \mathbf{x} = \mathbf{b}$, where:

\begin{equation}
\mathbf{A} = \begin{bmatrix}
-\frac{f}{d} & 0 & \frac{p_x}{d} & \frac{p_x p_y}{f} & -\frac{(p_x)^2 + f^2}{f} & p_y \\
0 & -\frac{f}{d} & \frac{p_y}{d} & \frac{(p_y)^2 + f^2}{f} & -\frac{p_x p_y}{f} & -p_x
\end{bmatrix},
\end{equation}

\begin{equation}
\mathbf{x} = \begin{bmatrix}
t_x \\ t_y \\ t_z \\ \theta_x \\ \theta_y \\ \theta_z
\end{bmatrix},
\quad \mathbf{b} = \begin{bmatrix}
F_x \\ F_y
\end{bmatrix}.
\end{equation}

With multiple flow vectors, the system becomes overdetermined and is solved using the least squares method:

\begin{equation}
\mathbf{x} = (\mathbf{A}^\top \mathbf{A})^{-1} \mathbf{A}^\top \mathbf{b}.
\end{equation}

This approach leverages the relationship between pixel displacements and camera pose, enabling a closed-form estimate of the relative pose parameters $\mathbf{t}$ and $\boldsymbol{\theta}$ from the flow and depths. With the exposure time given, we can then compute the instantaneous velocities $\mathbf{v}$ and $\boldsymbol{\omega}$ across exposure.

A convenient property of this linear least squares formulation is that this is differentiable, which allows us to train the network fully end-to-end with clear pose supervision. Therefore, our final end-to-end loss is:

\begin{equation}
    \mathcal{L}_2 = \lambda_R||R - h_p(\hat{R})||_2 + \lambda_t|| \mathbf{t} - h_p(\mathbf{\hat{t}})||_2 + \mathcal{L}_1
\end{equation}
\\
where we compute MSE on the rotations $R \in SO(3)$ and translations $\mathbf{t}$. $\lambda_R$ and  $\lambda_t$ help balance the loss, and $h_p$, similar to Equation \ref{eq:reorient_flow}, reorients the $\hat{R}$ or $\mathbf{\hat{t}}$ in such a way that is closest to the prediction.

Note that our current formulation assumes that the relative pose can be described with a single rigid transform, and we do not model for rolling shutter or non-uniform motion within exposure. However, this is still a reasonable assumption as we later show our results robust to real-world nonlinear blur and rolling shutter effects.



\subsection{Direction Disambiguation}
\label{sec:disambiguation}

While we can estimate the magnitude of the camera velocity directly, the direction has a $180^\circ$ ambiguity because the blurring process erases temporal cues. For instance, both leftward and rightward motion of the camera will result in the same horizontal blur traces. To address this, we introduce a simple and fast heuristic to resolve the directional ambiguity. Recall $\tau$ is the exposure time length and let $\Delta t_f$ be the inter-frame shutter start interval.
Since the motion is small between frames, we first approximate the flow field $\mathcal{F}_i$ to be linear. Then, for frame $I_i$, we linearly extrapolate $\mathcal{F}_i$ and apply the warping function $\Phi$ to $I_i$:

\begin{equation}
    \label{eq:flow_extrap}
    \mathcal{F}'_i = \frac{\Delta t_f}{\tau} \mathcal{F}_i
\end{equation}

\begin{equation}
    \label{eq:flow_warp}
    I'_i = \Phi(I_i; \mathcal{F}'_i)
\end{equation}

Because warping with $\mathcal{F}'_i$ is linear, warping the real frame is effectively warping each virtual frame and averaging them. Therefore, we can use $\mathcal{F}'_i$, the flow between the first virtual frames of $I_i$ and $I_{i+1}$, to warp the true frame $I_i$ (Eq. \ref{eq:flow_warp}). We use both the forward and backwards flow fields $\mathcal{F}_{fw}$ and $\mathcal{F}_{bw} = -\mathcal{F}_{fw}$ as $\mathcal{F}_i$ in Eq. \ref{eq:flow_extrap} to compute the extrapolated flow fields $\mathcal{F}'_{fw}$ and $\mathcal{F}'_{bw}$, respectively. Both $\mathcal{F}'_{fw}$ and $\mathcal{F}'_{bw}$ are used to compute the photometric error between the corresponding warped image $I'_i$ with the next true frame in the video sequence $I_{i+1}$:

\begin{equation}
    \mathcal{P}(I_1, I_2) = \frac{1}{H W} \sum_{u=0}^H \sum_{v=0}^W |I_1(u,v) - I_2(u,v)|.
\end{equation}

We additionally do this photometric error check with warping $\mathcal{F}'_i$ to $I_{i-1}$ and compute the sum of the photometric error between the two directions, and reorient based on the smaller photometric error.

\begin{equation}
\begin{split}
    e_{fw} = \mathcal{P}(I_{i+1}, I'_{i,fw}) + \mathcal{P}(I_{i-1}, I'_{i,bw})
    \\
    e_{bw} = \mathcal{P}(I_{i+1}, I'_{i,bw}) + \mathcal{P}(I_{i-1}, I'_{i,fw})
\end{split}
\end{equation}

\begin{equation}
\boldsymbol{\omega}, \mathbf{v} = 
\begin{cases}
      \boldsymbol{\omega}_{fw}, \mathbf{v}_{fw} & \text{if } e_{fw}  < e_{bw} \\
      \boldsymbol{\omega}_{bw}, \mathbf{v}_{bw} & \text{otherwise.}
\end{cases}
\end{equation}

In summary, we warp the image in the predicted and opposite flow directions, where we compare the photometric error to decide which direction is consistent with the video frames.
This heuristic enables us to quickly reorient the direction of the camera velocities in a consistent orientation.
\begin{figure}[tbp!]
  \centering
  \includegraphics[width=\columnwidth]{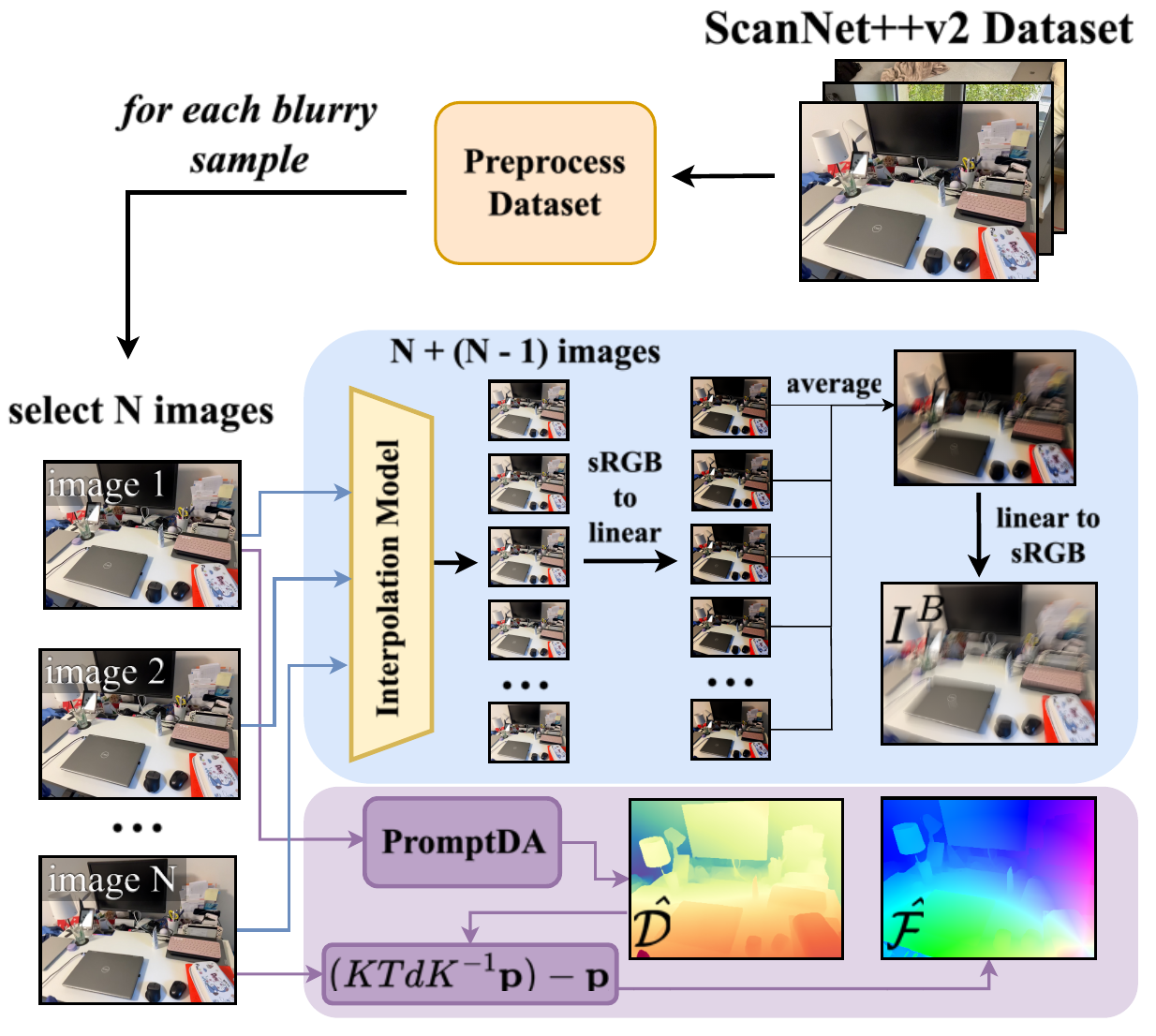}
  \caption{Overview for our synthetic dataset generation process. After preprocessing the dataset, we run selected frames through an interpolation network, which we use to synthesize our blurred image. We also take the first and last virtual frames to generate $\mathcal{\hat{D}}$, which is subsequently used for computing $\mathcal{\hat{F}}$.}
  \label{fig:blur_dataset}
\end{figure}

\section{Dataset Curation}
\label{sec:dataset}

\begin{table*}[h!]
  \centering
  \begin{tabular}{@{}c l cccc|c@{}}
    \toprule
    & \multirow{2}{*}{Method} & billiards & commonroom & dining & office & avg \\
    \cmidrule(lr){3-3} \cmidrule(lr){4-4} \cmidrule(lr){5-5} \cmidrule(lr){6-6} \cmidrule(lr){7-7}
    && $\omega_x$ / $\omega_y$ / $\omega_z$ 
           & $\omega_x$ / $\omega_y$ / $\omega_z$ 
           & $\omega_x$ / $\omega_y$ / $\omega_z$ 
           & $\omega_x$ / $\omega_y$ / $\omega_z$
           & $\omega_x$ / $\omega_y$ / $\omega_z$ \\
    \midrule
    \parbox[t]{1mm}{\multirow{4}{*}{\rotatebox[origin=c]{90}{MI}}} &
    COLMAP \cite{schoenberger2016sfm} \cite{schoenberger2016mvs}
    & $\times$
    & $\times$
    & $\times$
    & $\times$
    & $-$ \\
    & \ \ \rotatebox[origin=c]{180}{$\Lsh$} D+LG \cite{tyszkiewicz2020disk} \cite{lindenberger2023lightglue}
    & \underline{2.32} / \underline{1.94} / \textbf{1.50}
    & \underline{1.19} / \underline{1.61} / \textbf{0.91}
    & $\times$
    & \underline{3.06} / \underline{2.51} / \underline{3.93}
    & $-$ \\
    & MASt3R \cite{dust3r_cvpr24} \cite{mast3r_arxiv24}
    & 5.30 / 2.85 / 4.45
    & 3.70 / 3.75 / 3.26
    & \underline{2.36} / \underline{0.84} / \underline{1.67}
    & 4.78 / 3.03 / 6.21
    & \underline{4.04} / \underline{2.62} / \underline{3.90} \\
    & DROID-SLAM \cite{teed2021droid}
    & 5.39 / 3.33 / 5.31
    & 3.01 / 5.89 / 3.57
    & 2.92 / 1.20 / 1.98
    & 6.33 / 4.90 / 5.56
    & 4.41 / 3.83 / 4.10 \\
    \midrule
    \parbox[t]{1mm}{{\rotatebox[origin=c]{90}{SI}}}
    & Ours & \textbf{1.31} / \textbf{0.87} / \underline{1.60} & \textbf{0.93} / \textbf{0.88} / \underline{1.04} & \textbf{0.87} / \textbf{0.50} / \textbf{1.33} & \textbf{1.76} / \textbf{1.38} / \textbf{3.08} & \textbf{1.22} / \textbf{0.91} / \textbf{1.76} \\
    \midrule
    & \textcolor{gray}{Zero-Velocity baseline} & \textcolor{gray}{5.39 / 3.43 / 5.16} & \textcolor{gray}{3.95 / 4.50 / 2.81} & \textcolor{gray}{4.58 / 1.53 / 3.66} & \textcolor{gray}{5.43 / 3.19 / 6.99} & \textcolor{gray}{4.84 / 3.16 / 4.66} \\
    \bottomrule
  \end{tabular}
  \caption{RMSE for rotational velocities across each axis, in rad/s. We evaluate against multi-image (MI) and single-image (SI) methods. The best and second-best results are \textbf{bolded} and \underline{underlined}, respectively. The $\times$ represents a failure to reconstruct the poses.}  \label{tab:eval_angular}
\end{table*}

\begin{table*}[h!]
  \centering
  \begin{tabular}{@{}c l cccc|c@{}}
    \toprule
    & \multirow{2}{*}{Method} & billiards & commonroom & dining & office & avg \\
    \cmidrule(lr){3-3} \cmidrule(lr){4-4} \cmidrule(lr){5-5} \cmidrule(lr){6-6} \cmidrule(lr){7-7}
    && $v_x$ / $v_y$ / $v_z$ 
           & $v_x$ / $v_y$ / $v_z$  
           & $v_x$ / $v_y$ / $v_z$  
           & $v_x$ / $v_y$ / $v_z$ 
           & $v_x$ / $v_y$ / $v_z$  \\
    \midrule
    \parbox[t]{1mm}{\multirow{4}{*}{\rotatebox[origin=c]{90}{MI}}} &
    COLMAP \cite{schoenberger2016sfm} \cite{schoenberger2016mvs}
    & $\times$
    & $\times$
    & $\times$
    & $\times$
    & $-$ \\
    & \ \ \rotatebox[origin=c]{180}{$\Lsh$} D+LG \cite{tyszkiewicz2020disk} \cite{lindenberger2023lightglue}
    & 3.11 / 2.06 / 2.14
    & 1.70 / 2.09 / 2.20
    & $\times$
    & 1.81 / 2.45 / 1.52
    & $-$ \\
    & MASt3R \cite{dust3r_cvpr24} \cite{mast3r_arxiv24}
    & \underline{2.59} / \underline{1.50} / 3.52
    & \underline{1.28} / 1.94 / 2.49
    & \textbf{0.82} / \textbf{0.55} / 1.22
    & \underline{1.72} / \underline{2.16} / 1.46
    & \underline{1.60} / \underline{1.54} / 2.17 \\
    & DROID-SLAM \cite{teed2021droid}
    & 3.28 / 2.00 / \underline{2.65}
    & 2.23 / \underline{1.47} / \underline{1.25}
    & 1.44 / 0.65 / \underline{0.91}
    & 1.98 / 2.40 / \underline{1.23}
    & 2.23 / 1.63 / \underline{1.51} \\
    \midrule
    \parbox[t]{1mm}{{\rotatebox[origin=c]{90}{SI}}}
    & Ours & \textbf{1.36} / \textbf{1.17} / \textbf{1.05} & \textbf{1.00} / \textbf{0.80} / \textbf{0.71} & \underline{0.95} / \underline{0.61} / \textbf{0.81} & \textbf{1.12} / \textbf{1.52} / \textbf{1.12} & \textbf{1.11} / \textbf{1.03} / \textbf{0.92} \\
    \midrule
    & \textcolor{gray}{Zero-Velocity baseline} & \textcolor{gray}{2.80 / 1.66 / 1.41} & \textcolor{gray}{1.98 / 1.48 / 1.26} & \textcolor{gray}{1.30 / 1.05 / 1.20} & \textcolor{gray}{1.94 / 2.26 / 1.07} & \textcolor{gray}{2.01 / 1.61 / 1.24} \\
    \bottomrule
  \end{tabular}
  \caption{RMSE for translational velocities across each axis, in m/s. We evaluate both multi-image (MI) and single-image (SI) methods. The best and second-best results are \textbf{bolded} and \underline{underlined}, respectively. The $\times$ represents a failure to reconstruct the poses.}
  \label{tab:eval_translate}
\end{table*}

\subsection{Synthetic Dataset}

To train our model, we require a dataset of blurred images with corresponding ground-truth flows, depths, and relative poses across exposure. Since no public dataset meets these requirements, we create our own by adapting a subset of ScanNet++v2 \cite{yeshwanthliu2023scannetpp}. Figure \ref{fig:blur_dataset} shows our data pipeline. ScanNet++v2 consists of about 1000 iPhone videos of indoor scenes with COLMAP \cite{schoenberger2016sfm} \cite{schoenberger2016mvs} and ARKit-captured poses. We are interested in obtaining relative poses across short time spans (about 0.01 s) and robust to fast-motion, so we opt to use filtered ARKit poses as our ground truth.

We scope this paper to the indoor domain to highlight scenes more susceptible to blur and noise. We do not train on outdoor scenes, so we do not expect our performance to be as accurate outdoors as it is indoors. However, our data pipeline is flexible in that it can easily use any video to generate motion-blurred images, therefore allowing us to obtain images in different domains when desired.

\paragraph{Synthesizing blur.} We now show how we create a motion blurred image $I^B$ for training. For a selected RGB image $I^\nu_i$ in a video, we obtain the following $N$ images to have in total $N+1$ images $\{I^\nu_i, ..., I^\nu_{i+N}\}$. We then use an off-the-shelf frame interpolation model RIFE \cite{huang_rife_2022} to interpolate between each consecutive frame, resulting in a total of $2N + 1$ virtual frames $\mathcal{I} = \{I^\nu_i, I^\nu_{i + \frac{1}{2}}, \dots, I^\nu_{i + N}\}$. We follow Equation \ref{eq:approx_blur} and first convert $\mathcal{I}$ in linear space $g^{-1}(\cdot)$ before averaging all images and converting back to sRGB space to create $I^B$. Even at 60 FPS, frames are sampled too slowly to be averaged for rendering realistic motion blur, so we further interpolate between real-world frames to generate more realistic motion trajectories and facilitate better generalization to in-the-wild motion blur. We curate images with $N=3$ for small motion and $N=10$ for large motion. We emphasize that while we average multiple frames to generate synthetic motion-blurred images, during training/inference our model only ever receives one image at a time.

\paragraph{Obtaining depth and flows.} We only use the first and last virtual frames $I^\nu_1$ and $I^\nu_N$ for obtaining the ground-truth flow field. We use the low-resolution ARKit depth and $I^\nu_1$ in PromptDA \cite{lin2024promptda} to obtain our dense, high-resolution depth map $\mathcal{\hat{D}}$ as ground truth. For all pixels, we then perform
$
    \mathbf{p'} = K T d K^{-1} \mathbf{p},
$
which uses $d \in \mathcal{\hat{D}}$ to backproject all pixel points in $I^\nu_1$ and subsequently project them into $I^\nu_N$. Afterwards, we retrieve the flow field $\mathcal{\hat{F}}$ with the displacements of the projected points as expressed in Equation \ref{eq:flow_def}.

In total, we synthesize about 120k training and 1.2k validation samples across 150 ScanNet++v2 sequences, with corresponding blurred images, depth maps and flow fields.

\subsection{Real-world Dataset}

As our method, including the velocity solver, is fully-differentiable, we can train it in an end-to-end manner. This enables us to additionally train on real-world motion blur to further close the reality gap between our synthetic dataset and in-the-wild images. We collect 10k real-world motion-blurred images with corresponding ARKit poses, IMU measurements, and exposure times across about 30 scenes. We follow Liu et al. \cite{liu2008cvpr_blur-detect} and perform the Fast Fourier Transform to identify blurred images, where images within a threshold are blurry enough to be used. We also add images without blur to enhance the robustness to still frames.
\begin{figure*}[h!]
  \centering
\includegraphics[width=1.0\linewidth]{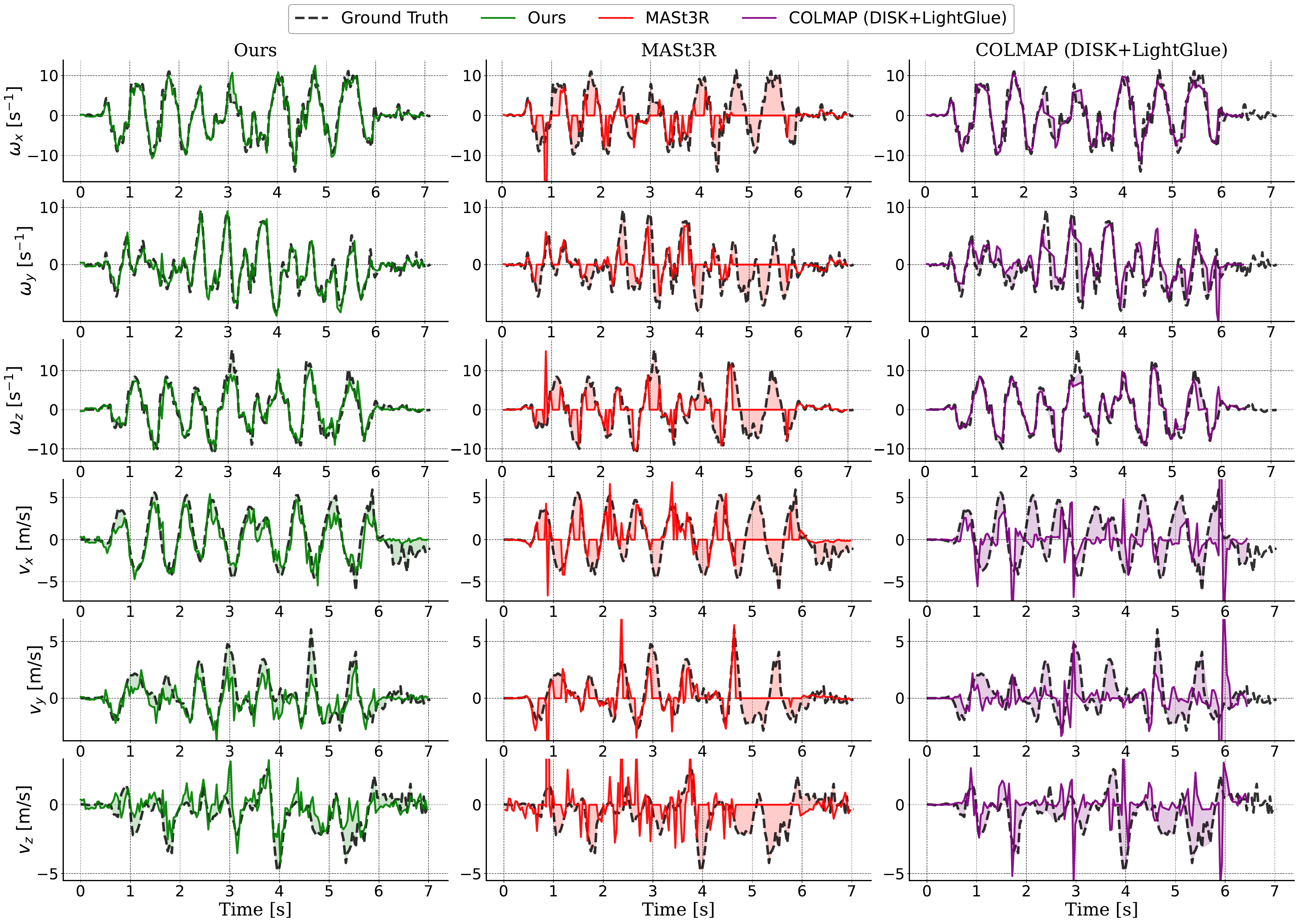}
  \caption{Visualization of the predicted velocities for the billiards sequence using our method, MASt3R and COLMAP (w/ DISK+LightGlue). The shaded area under the curve shows the error between the predicted velocity and GT velocity. Our translations and rotations are significantly better than MASt3R. While COLMAP with DISK + LightGlue feature matching does well on rotations, our method significantly outperforms it on translations. }
  \label{fig:vel_preds}
\end{figure*}

\section{Experiments}
\label{sec:experiments}

\paragraph{Training.} Our model is trained in three stages. We first train the flow and depth decoders without pose supervision on our synthetic dataset with a batch size of 32. Pose supervision is added once the depths and flow estimates are reasonable, and we train poses, flow and depths with a batch size of 8 for 300k steps. Finally, we finetune the model on real-world collected images for 10k steps. We train our method using Adam \cite{kingma2017adam} on an Nvidia RTX 3090 GPU.

\paragraph{Evaluation.}

To assess the accuracy, we evaluate the rotational velocities $\boldsymbol{\omega}$ and translational velocities $\mathbf{v}$ in four unseen, real-world videos.
These test sequences explicitly have different domains from those in training and have varying exposure times to test its generalization.
We do not evaluate on existing motion-blurred benchmarks such as GoPro \cite{Nah_2017_CVPR} since neither poses nor velocities are provided, and computing poses using COLMAP and MASt3R were inaccurate due to small baselines.
For each test sequence, we get the ARKit-estimated camera poses, IMU measurements, and the exposure times for each image. We directly compare $\boldsymbol{\omega}$ predictions to the gyroscope measurements and $\mathbf{v}$ predictions to a finite-difference velocity approximation from the ARKit translations. We use these translations as our ground-truth since ARKit's visual-inertial method tightly fuses IMU measurements and is robust even with poor visual tracks. We also only use the computed ARKit instantaneous body-frame velocity, mitigating any major drift. We do not use COLMAP as we found it unreliable in these scenarios (see Tab. \ref{tab:eval_angular} and \ref{tab:eval_translate}). We measure the RMSE across each axis for $\boldsymbol{\omega}$ and $\mathbf{v}$. 

\paragraph{Baselines.}
No existing methods estimate camera motion from a single blurred image, so we instead use SfM/SLAM methods and estimate the relative pose (and hence velocity) between frames using a centered finite-difference approximation. We evaluate against the following baselines:

\begin{itemize}
    \item \textit{COLMAP} \cite{schoenberger2016sfm} \cite{schoenberger2016mvs} is an incremental-SfM method that is considered the de-facto method for obtaining camera poses. Because standard SfM is unreliable with small triangulation angles, we run COLMAP with all sequence images. After estimating the trajectory, we scale it using the ground-truth ARKit poses. We use \texttt{hloc} \cite{sarlin2019coarse} to run COLMAP with SIFT \cite{Lowe2004DistinctiveIF} + AdaLAM \cite{cavalli2020handcrafted} as well as a learned baseline with DISK \cite{tyszkiewicz2020disk} + LightGlue \cite{lindenberger2023lightglue}.
    \item \textit{MASt3R} \cite{dust3r_cvpr24} \cite{mast3r_arxiv24} is a recent learning-based 3D vision method that regresses the pointmaps between a pair of images. MASt3R is the current SOTA in retrieving metric-scale camera poses, particularly with sparse views with little-to-no overlap. To better compare to our method, we evaluate MASt3R on pairwise images two frames apart.
    \item \textit{DROID-SLAM} \cite{teed2021droid} is a learning-based SLAM method that uses RAFT \cite{teed_raft_2020} to estimate the optical flow across frames to help compute the camera poses. To obtain the camera poses for each frame, we set the filter and keyframe thresholds to 0.
\end{itemize}

We also provide the zero-velocity baseline representing a stationary camera, which serves as a reasonable prior for most trajectories.
Note that all baselines require two or more images to estimate the velocity, whereas our model \textit{only ever sees a single image}. We therefore tackle a more difficult problem, and our formulation more closely resembles the true camera state.

\begin{figure}[tb]
  \centering
\includegraphics[width=1.0\linewidth]{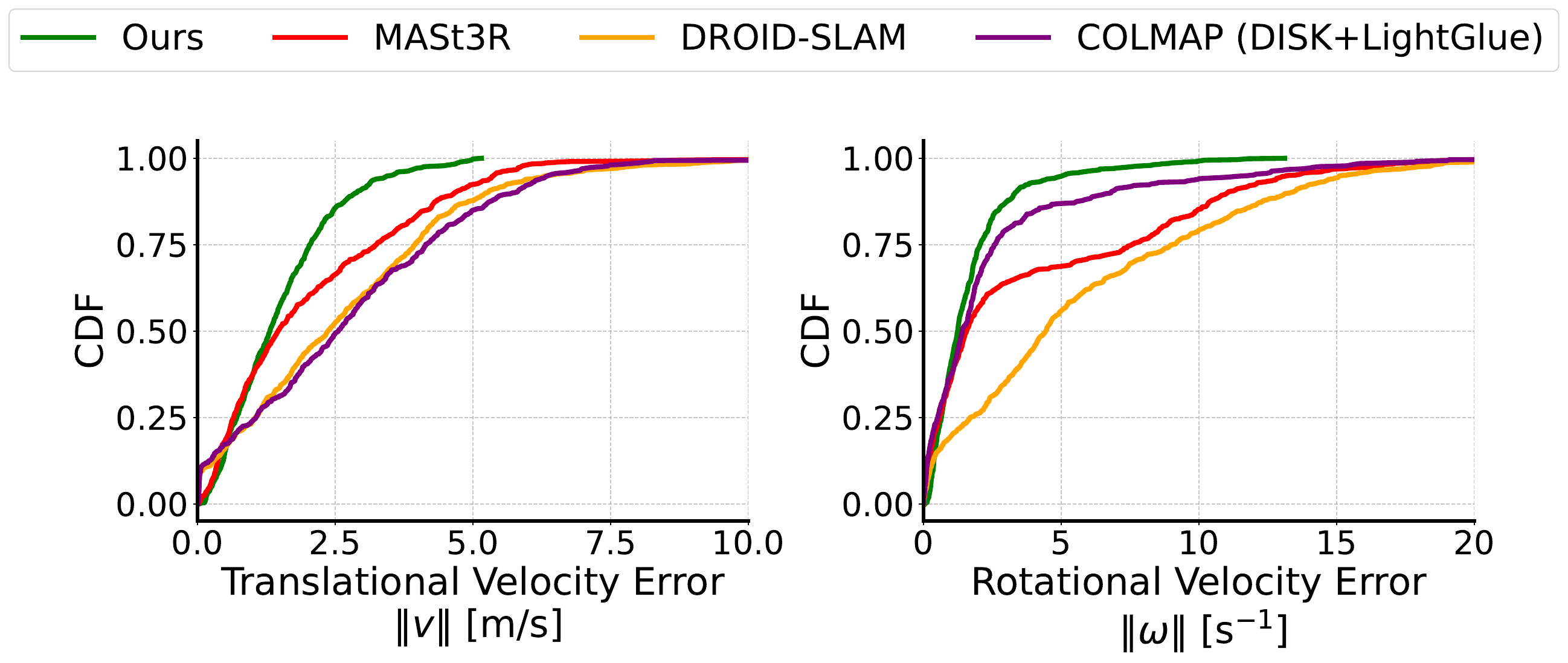}
  \caption{Error CDFs across all test sequences, such that the left and right plot show the distribution of translational and rotational errors, respectively. Curves closer to the top-left corner are better.}
  \label{fig:error_cdfs}
\end{figure}

\begin{figure}[tb]
  \centering
  \begin{subfigure}{0.35\columnwidth}
    \includegraphics[width=1.0\linewidth]{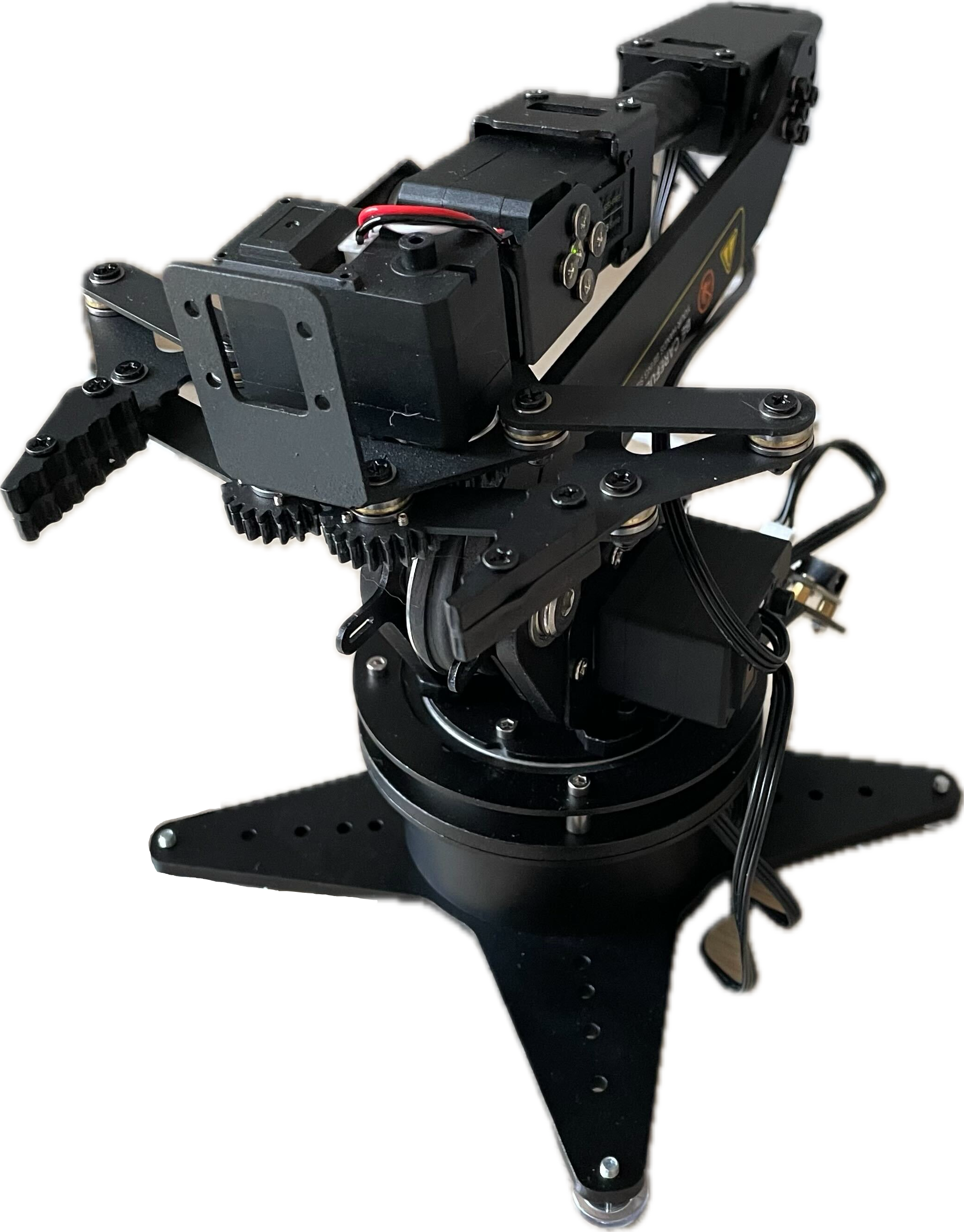}
  \end{subfigure}
  \hfill
  \begin{subfigure}{0.62\columnwidth}
    \includegraphics[width=1.0\linewidth]{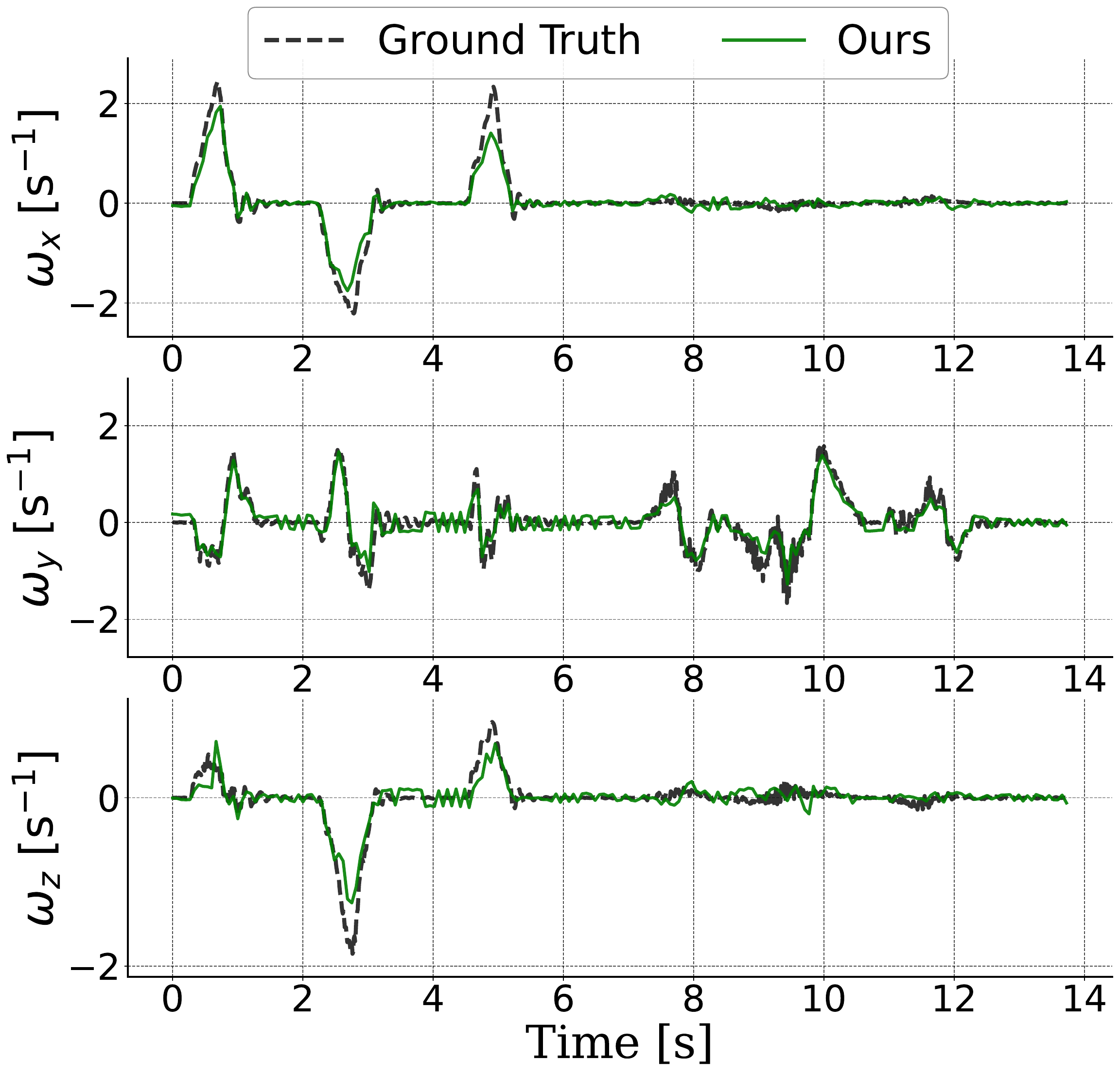}
  \end{subfigure}
  \caption{Real-world application example of our method using a camera attached to a fast-moving RoArm-M1 robot arm platform. (a) The robot arm used for recording. (b) The predicted and GT velocity time series for the camera attached to the end-effector.}
  \label{fig:robot_arm}
\end{figure}

\begin{figure}[tb]
    \centering
    \includegraphics[width=\columnwidth]{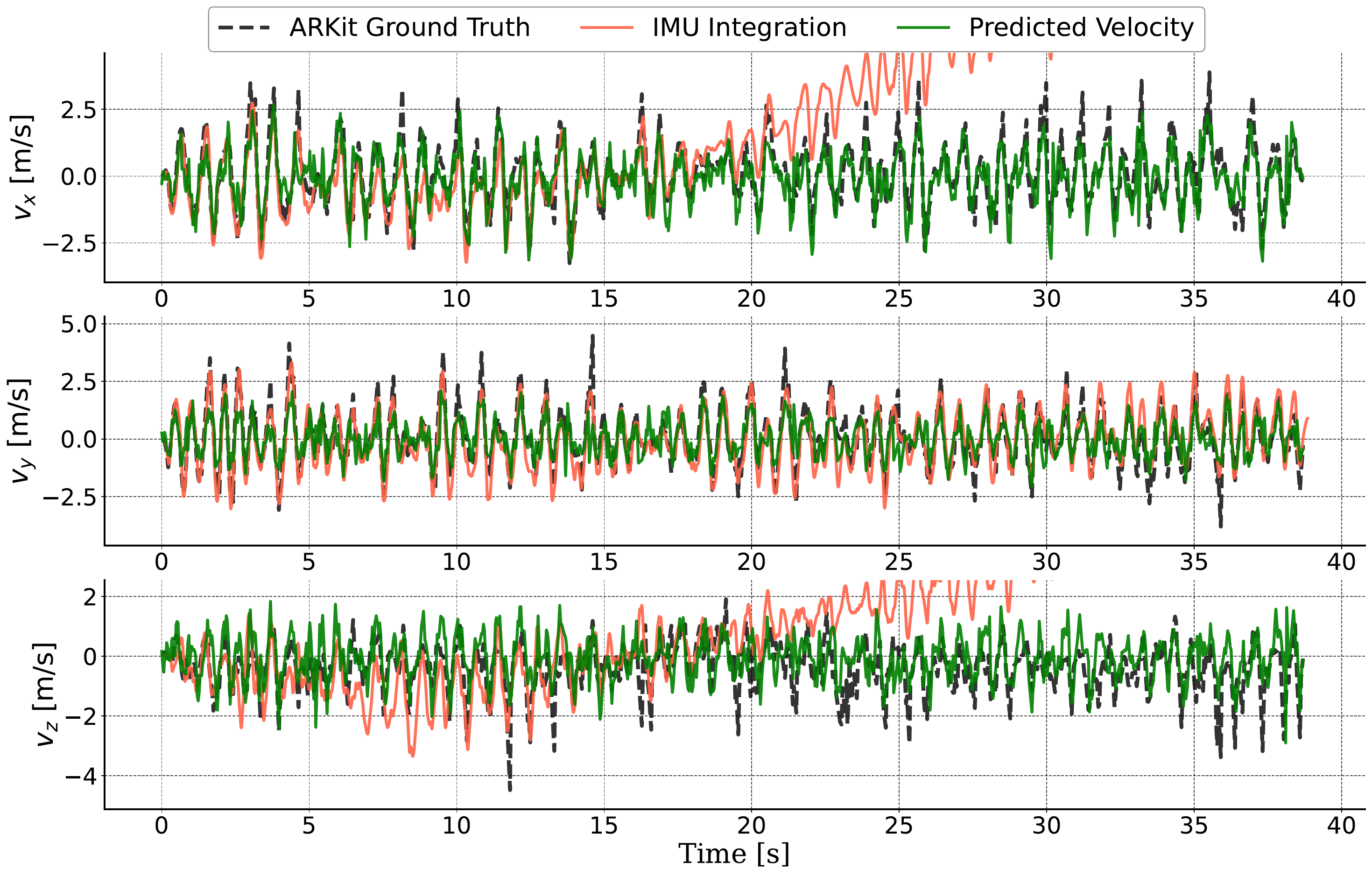}
    \caption{Our method vs using IMU integration. The IMU velocity estimate is accurate for a few seconds but drifts. Our method provides accurate and drift-free estimates throughout the sequence.}
    \label{fig:imu_comparison}
\end{figure}

\begin{table}[tb]
  \centering
  \begin{tabular}{@{}l|c@{}}
    \toprule
    Method & \textcolor{black}{Average FPS} \\
    \midrule
    COLMAP (D+LG) & 0.95 \\
    MASt3R & 2.83 \\
    DROID-SLAM & 4.34 \\
    Ours & \textbf{30.17} \\
    \bottomrule
  \end{tabular}
  \caption{Average method FPS across all test sequences, computed by number of processed frames over runtime. All methods were also averaged across 3 runs for each sequence. For COLMAP, we only compute across sequences that did not fail.}
  \label{tab:eval_runtime}
\end{table}

\paragraph{Results.}

We report the rotational and translational RMSE in Tables \ref{tab:eval_angular} and \ref{tab:eval_translate}. While the recorded video sequences suffer from severe motion blur (see supplementary for examples), our method is able to handle the motion robustly. Figure \ref{fig:error_cdfs} highlights the overall robustness of our method, with the error CDFs demonstrating that our method suffers from noticably less errors than the baselines. In these sequences, the baseline methods suffer from major outliers and zero-prediction failures, observed in Figure \ref{fig:vel_preds}. Overall, we obtain an average of 31\% reduction of RMSE from the second-best result in rotational velocities and a 24\% reduction in translational velocities.

We also test our method in a robotics application by estimating the gripper velocity on a robot arm. We place the camera on the robot pictured in Figure \ref{fig:robot_arm}(a) and predict the gripper velocity solely from its video. The motion-blurred images give a salient enough signal to accurately estimate the movement of the arm as seen in Figure \ref{fig:robot_arm}(b).

In Figure \ref{fig:imu_comparison} we compare our method with using an actual IMU. We first estimate the IMU bias from a stationary calibration video, followed by estimating the initial pose of the camera to properly align with the gravity vector. We then preintegrate the IMU measurements following \cite{forster2016manifold} using GTSAM \cite{gtsam} and compare the integrated velocities to our predictions. Figure \ref{fig:imu_comparison} shows the drift that occurs in $v_x$ and $v_z$ after only 20 seconds, whereas our method is reliably drift-free for the entire video. While we are comparing our direct translational velocity estimates to the IMU's preintegrated acceleration, we argue that this is advantageous for state estimation since we bypass the drift-accruing integration. Therefore, we can see how our method can provide an \textit{even better} motion estimate than an IMU.

We lastly measure each method's average FPS (Table \ref{tab:eval_runtime}). Even when running on every frame, our method can run in real-time at 30 FPS, with inference and direction disambiguation taking 0.03 seconds on an RTX 3090 GPU.
\section{Conclusion}
\label{sec:conclusion}

In this paper we have proposed a method that can estimate the instantaneous camera velocity from individual motion-blurred images -- producing inertial-like measurements without an IMU. Our evaluations indicate that not only do we obtain more accurate and robust results compared to an actual IMU and other vision-based solutions, but we also do so significantly more efficiently. We hope that this motivates the community to explore further how traditionally unwanted or overlooked cues can be used to enhance the performance of vision tasks.
{
    \small
    \bibliographystyle{ieeenat_fullname}
    \bibliography{main}
}


\end{document}